\pdfoutput=1

\documentclass[11pt]{article}
\usepackage[hyperref]{acl}

\usepackage{times}
\usepackage{latexsym}
\usepackage[T1]{fontenc}
\usepackage[utf8]{inputenc}
\usepackage{microtype}
\usepackage{inconsolata}
\usepackage{amsmath}
\usepackage{amssymb}
\usepackage{mathtools}
\usepackage{amsthm}
\usepackage{xspace}
\usepackage{graphicx}
\usepackage{multirow}
\newcommand{\model}{\textsc{Pixar}\xspace}
\newcommand{\pixel}{\textsc{Pixel}\xspace}
\usepackage{booktabs}
\newcommand{\im}{\boldsymbol{x}}

\usepackage{algorithm}
\usepackage{algpseudocode}
\usepackage{bm}
\usepackage{algcompatible}
\usepackage{microtype}
\usepackage{graphicx}
\usepackage{booktabs} 
\usepackage{subcaption}
\usepackage[font=normalsize]{caption}
\usepackage{hyperref}
\usepackage[capitalize,noabbrev]{cleveref}
\hypersetup{
colorlinks=true,linkcolor=teal,citecolor=magenta
}

\title{\model: Auto-Regressive
Language Modeling in Pixel Space}

\author{
    Yintao Tai$^*$ \\ 
    University of Edinburgh \\ 
    \And 
    Xiyang Liao$^*$ \\ 
    University of Edinburgh \\ 
    \And 
    Alessandro Suglia$^\dagger$ \\ 
    Heriot-Watt University \\ 
    \AND 
    Antonio Vergari$^\dagger$ \\
    University of Edinburgh\\
}

\begin{document}
\maketitle

\def\thefootnote{*}\footnotetext{Equal contribution}\def\thefootnote{\arabic{footnote}}
\def\thefootnote{$\dagger$}\footnotetext{Joint supervision,  correspondence to: Alessandro Suglia <A.Suglia@hw.ac.uk>, Antonio Vergari <avergari@ed.ac.uk>}\def\thefootnote{\arabic{footnote}}

\begin{abstract}

Recent work showed the possibility of building open-vocabulary large language models (LLMs) that directly operate on pixel representations. These models are implemented as autoencoders that reconstruct masked patches of rendered text.
However, these pixel-based LLMs are limited to discriminative tasks (e.g., classification) and, similar to BERT, cannot be used to \textit{generate text}.
Therefore, they cannot be used for generative tasks such as free-form question answering.
In this work, we introduce \model, the first pixel-based autoregressive LLM that performs text generation.
Consisting of only a decoder, \model can perform free-form generative tasks 
while keeping the number of parameters on par with previous encoder-decoder models.
Furthermore, we highlight the challenges of generating text as non-noisy images and show this is due to using a maximum likelihood objective.
To overcome this problem, we propose an adversarial pretraining stage that improves the readability and accuracy of \model by 8.1 on LAMBADA and 8.5 on bAbI--- making it comparable to GPT-2 on text generation tasks.
This paves the way to build open-vocabulary LLMs that operate on perceptual input only and calls into question the necessity of the usual symbolic input representation, i.e., text as (sub)tokens.

\end{abstract}

\begin{figure}[t]
\centering
\includegraphics[width=0.99\columnwidth]{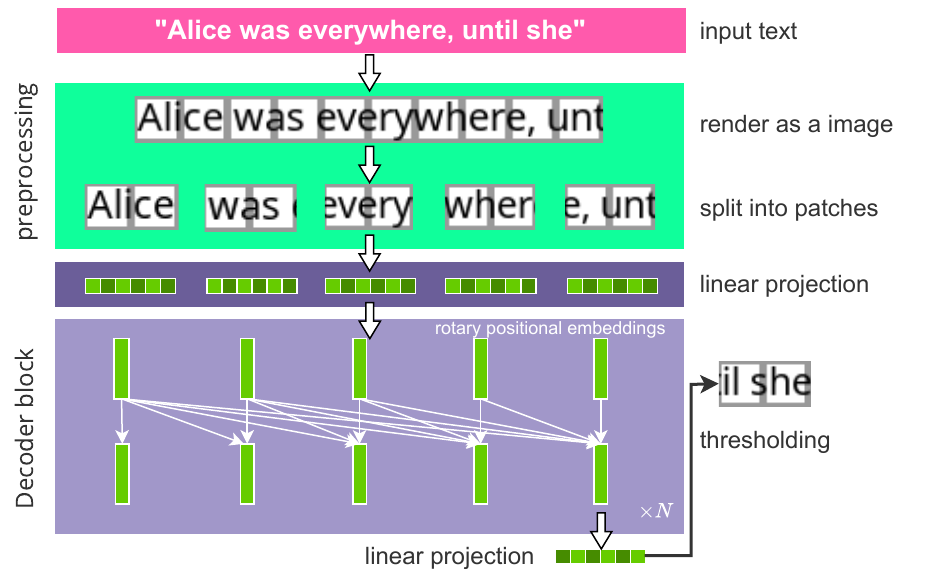}
\caption{\textbf{\model is the first generative language model operating on pixels only.} \model accepts texts as images and also generates texts in image patches autoregressively, a challenging task. 
}
\label{fig:pixararch}
\end{figure}

\section{Introduction}
\label{sec:introduction}

In natural language processing (NLP) pipelines, 
tokenizers are an essential ingredient used to divide the raw text into a sequence of sub-units, such as sub-words \citep{wordpiece}, characters \citep{macbert}, sentence pieces \citep{sentencepiece}, and bytes \citep{bpe}. 
Normally, NLP models represent these sub-units as \textit{symbols} within an ID-based categorical vocabulary. 

This categorical representation highlights the following weaknesses of traditional NLP systems.
First, to encode each item in the vocabulary, NLP systems have to allocate an embedding matrix 
that grows linearly with vocabulary size. 
Modern large language models (LLMs) allocate millions of parameters just for this.\footnote{The embedding matrix of Vicuna-7B ($\sim$130M) is comparable to the size of \texttt{bert-base-uncased} ($\sim$109M).}
Additionally, fixing a vocabulary a priori can lead to performance degradation due to unseen out-of-vocabulary (OOV) words \citep{challenge}.
Tokenizers with smaller granularities, such as characters and bytes, can alleviate the OOV issue but are still brittle as they can suffer from orthographic attacks \citep{visualattack}.
On the other hand, humans are incredibly robust to a variety of text permutations~\cite{rayner2006raeding} because they leverage the graphical information in text \citep{chinesebert}.

To tackle these problems, \citet{pixel} proposed \pixel, a pixel-based LLM that treats text as images.
Pixel-based embeddings remove the need for a finite vocabulary and keep the visual information of text, \textit{questioning whether we need symbolic representations of text as input at all}, or if an LLM can learn symbols implicitly.
\pixel achieved comparable performance with BERT \cite{bert} in a range of downstream classification and regression NLP tasks while being robust to character-level visual attacks \citep{visualattack}.
However, because of its close architectural similarities with BERT, 
\pixel cannot deal with free-form generative tasks, such as generative question answering \cite{future_token}.

To fill this gap, we present \model\footnote{
Code is publicly available at \href{https://github.com/april-tools/pixar}{https://github.com/april-tools/pixar}.
},
the first pixel-based autoregressive LLM that can generate short sequences of text as images.
\model is to GPT-like architectures as \pixel is to BERT-like architectures:
it consists of a Transformer decoder \citep{gpt2} that autoregressively generates text image patches as output. 
\textit{Generating new text as pixels starting from pixels only} is, however, more challenging than selecting symbolic tokens from a vocabulary (as GPT-like models) or reconstructing masked image patches (as in \pixel).
This is because the model has to learn to generate longer sequences of pixels.
To this end, we introduce a two-stage pretraining strategy for \model.
First, following previous work on autoregressive LLMs \citep{gpt2} and image generation models~\cite{chen2020imagegpt}, \model is trained by reconstructing the next patch of pixels derived from a large-scale corpus of rendered text using teacher-forcing. 
This maximum-likelihood approach, however, can generate image patches containing noisy text.
To mitigate this problem, we proposed a second pretraining stage, where \model is trained with an additional adversarial loss.

Our experiments in Section \ref{sec:experiments} show that 200 steps of stage 2 pretraining improve the readability of generated text significantly and achieve comparable performance with GPT-2 \cite{gpt2} on open-answer short generative tasks such as bAbI \citep{babi} and LAMBADA \citep{lambada}.
Additionally, \model achieves better performance than \pixel on the GLUE benchmark~\cite{glue} while using a computational budget and a number of model parameters equivalent to the encoder part of \pixel.

\section{Beyond token-based LLMs}
\label{sec:background}

The idea of using pixel-based representations of text has been applied in various NLP tasks.
For instance, \citet{chinese1} used a CNN-based block to extract character-level visual representations of Chinese writing and 
similarly, \citet{superchar} for text classification.
Graphical features of Chinese characters are a good example where canonical symbolic tokenizers miss important information, as highlighted in 
\citet{tianzige} exploiting the Tianzige feature of Chinese characters or
\citet{mix} and ChineseBERT \citep{chinesebert} integrating character-level visual features into embedding vectors for BERT-like models.

Sub-word tokenization as in character-level or byte-level tokens is possible and partially solves some issues.\footnote{We discuss more related works in \cref{app:byte-level}.}
However, these approaches still struggle with a medium that is inherently visual. This is the case if we consider certain writing systems that are (partially) logographic such as in Chinese, or 
when text contains emojis.

Pixel-based representations mitigate these issues.
\citet{vtrans} built machine translation models using visual information as input. 
However, their output layers still rely on embeddings over a fixed vocabulary.
Inspired by \citet{vtrans}, 
\pixel \citep{pixel} pretrained a Masked AutoEncoder (MAE) \citep{mae} with a large corpus of rendered text using a masked reconstruction objective. 
This can be considered the first \textit{purely pixel-based LLM} that can be applied to a variety of downstream tasks such as POS tagging and extractive question answering. This approach was extended by \citet{salesky2023multilingual} to deal with multiple languages as well.
Moreover, visual text representation can be utilized in multimodal models to build a unified representation for both image and text modalities. Specifically, CLIPPO \citep{tschannen2023clippo} replaces the ID-based text encoder of CLIP \citep{clip} with a single pixel-based encoder to process both regular images and rendered text for visual QA.
Due to the encoder-only architecture of \pixel and CLIPPO, they still cannot be used to generate text.
GlyphDiffusion \citep{li2023glyphdiffusion} made attempts to use a diffusion model to generate new texts as images from noise and conditioned on encoded text features. 
However, its encoder still relies on symbolic embeddings.
On top of the benefits of purely pixel-based LLMs, we are interested in \textit{understanding if a deep learning model can learn symbolic representation from perceptual information only}, a very challenging task, as discussed next.

\section{\model}
\label{sec:method}

In contrast with the aforementioned approaches, we design \model as a PIXel-based AutoRegressive LLM whose inputs and outputs are only pixels representing textual information.
\model abandons the MAE architecture of \pixel and instead applies the design choices of other generative LLMs like GPT-2 and LLaMA-2~\cite{llama,llama2}. 
\model is pretrained to predict the next patches of pixels, conditioning only on the previous patches of rendered text. 
While this allows \model to tackle free-form generative tasks, it presents several challenges.
We start by reviewing its architecture.

\subsection{Model Architecture}
\label{subsec:architecture}

We design \model as a decoder-only model with a stack of $N=12$ Transformer layers.
Specifically, we extend the Transformer \cite{transformers} by applying
some of the improvements introduced with LLaMA-2:
pre-normalization using RMSNorm \cite{rmsnorm}, SwiGLU activation functions \cite{swiglue}, and rotary positional embeddings \cite{rope}. 
Each Transformer layer in \model generates $h^{\text{out}} \in \mathbb{R}^d$ hidden states as output. 
Further details are in Appendix \ref{app:pretraindetail}.

Similarly to \pixel, we represent the input text as a single (long) image containing several non-overlapping patches (\cref{fig:pixararch}).
Following \citet{mae}, each image patch $\boldsymbol{x} \in \mathbb{R}^{H \times W \times C}$---where $H$ and $W$ are the patch height and width and $C$ the number of channels---is then flattened into a vector, and then projected into a hidden embedding $h^{\text{in}} \in \mathbb{R}^d$. The resulting sequence of patch embeddings $\{h^{\text{in}}_1, \dots, h^{\text{in}}_{\text{eos}}\}$ is used as input to the Transformer backbone.
We experiment with both RGB images (i.e., $\boldsymbol{x} \in [0, 1]^{H \times W \times 3}$) and binary images ($C=1$ and $\boldsymbol{x} \in \{0,1\}^{H \times W}$), finding the latter to provide equivalently good downstream performance while simplifying the learning problem (\cref{sec:experiments}).

A linear layer after the last Transformer layer projects the embedding $h^{\text{out}}_N$ 
back to pixel space as a vector $\tilde{\boldsymbol{x}}$ representing a linearized $H\times W\times C$ patch (with $C=1$ for binary images). 
We interpret $\tilde{\boldsymbol{x}}$ in the following way.
For text rendered as binary images, 
$\tilde{\boldsymbol{x}}$ are the logits that then are squashed by an element-wise sigmoid with temperature $T$ ($T=1)$.
To recover a hard binary vector from these probabilities, we apply a threshold $\theta$ ($\theta = 0.5$).

When dealing with RGB images, we first clip $\boldsymbol{\tilde x}$ element-wise to be within  $[0, 1]$, then linearly map each channel to \{0-255\} to construct RGB patches. 

\subsection{Stage 1 training: MLE}
We train \model by maximum likelihood estimation (MLE) by minimizing the negative log-likelihood of $L$ ground truth pixel patches $\boldsymbol{x}_{i:i+L-1}$ conditioned on a sequence of observed (gold) patches $\boldsymbol{x}_{1:i-1}$ which is known as ``teacher forcing"~\cite{williams1989learning}. 
We assume every pixel in $\boldsymbol{x}_{i:i+L}$ to be conditionally independent given the last layer embedding $h^{\text{out}}_N$.
This yields to minimize a reconstruction loss $\mathcal{L}_{\mathsf{rec}}$ over $\boldsymbol{x}_{i:i+L-1}$ that ends up being
the usual pixel-wise binary cross-entropy loss for binary images, and the MSE loss for RGB images under our assumptions \citep{kingma2013auto,ghosh2020variational}.

\begin{figure}[!t]
\centering
\includegraphics[width=0.5\textwidth]{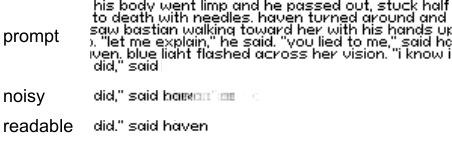}
\caption{\textbf{A second stage adversarial training can improve the readability of text generated} by \model when compared to noisy patches generated by MLE only. 
The prompt is from the LAMBADA test set, rendered as a binary image.
}
\label{fig:readability}
\end{figure}

This sequential prediction task is quite challenging, as we will need to predict $H\times W\times C\times L$ variables.
Having text rendered as binary images simplifies learning, but we observed that \model can easily get stuck in local optima that yield noisy generation: especially for $L>1$, see \cref{fig:readability}.
This is somehow expected and linked to the tendency of MLE to put probability mass among possible modes \citep{theis2015note} i.e., readable patch configurations in our case.
In \cref{sec:readability} we introduce a way to quantify ``readability'' of the generated patches, while we discuss next a practical solution to avoid the pitfalls of MLE training.

\subsection{Stage 2 training: Adversarial}
\label{sec:readability}

Our solution to the noisy generated patches is an adversarial loss to be used in conjunction with $\mathcal{L}_{\mathsf{rec}}$ in a second stage training.
We name it patch-wise context-aware adversarial (PCAA) loss, 
and noticed 
that only 200 steps of minimizing it 
in  
our experiments (\cref{sec:gentask}) can greatly boost the readability and generation performance of \model.

\begin{figure*}[!t]
    \centering
    \small
\setlength{\tabcolsep}{2pt}
    \begin{tabular}{ccccc}
    \rotatebox{90}{
         \textsc{bAbI}
         }
         \rotatebox{90}{
         prompt
         }& 
         \includegraphics[width=.465\textwidth]{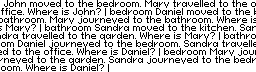}
         &
         \rotatebox{90}{
         generated
         }
         &
         \includegraphics[width=.465\textwidth]{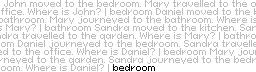}\\
         \rotatebox{90}{
         \textsc{LAMBADA}
         }
         \rotatebox{90}{
         prompt
         }& 
         \includegraphics[width=.465\textwidth]{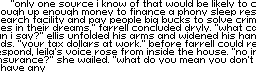}
         &
         \rotatebox{90}{
         generated
         }
         &
         \includegraphics[width=.465\textwidth]{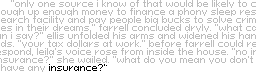}
         \end{tabular}
    \caption{\textbf{\model can generate readable and correct texts} according to the prompt (darker) on both bAbI (top) and LAMBADA (bottom). We folded images into rectangles for better visibility. 
    }
    
    \label{fig:gensample}
\end{figure*}

\textbf{Context-aware adversarial loss.}
Following previous work on GANs~\cite{gan}, an adversarial training regime consists of a generator and a discriminator. We consider \model as our generator. When designing our discriminator, we tried to follow previous work that uses patch-wise discriminators \cite{taming, stablediffu}. However, in preliminary experiments, we find that these solutions lead to catastrophic mode collapse because they only consider the predicted patch as input and ignore previously predicted patches, i.e., the ``context".

Our solution is the PCAA loss:
$\mathcal{L}_{\textsf{PCAA}} = \mathbb{E}_{\tilde{\im}_{i:i+L-1}}[-\log(D(\tilde{\boldsymbol{x}}_{i:i+L-1} | \im_{1:i-1}))]$, which represents how much the generator can ``fool" the discriminator a generated patch $\tilde{\im}_i$ is real.
To compute the PCAA loss, we use a context-aware discriminator implemented by appending a patch-wise classification head to a copy of the pretrained \model (stage 1).
Then, for a sequence, the discriminator is trained to decide whether a patch $\im_i$ is real or fake (i.e., ``readable text'' vs ``noisy'' one) given real previous patches $\im_{1:i-1}$.

Due to the computational overhead of the Transformer layers in \model, we developed a patch-sampling algorithm to effectively calculate the PCAA loss.
For a sequence of patches, we first generate fake patches and calculate the reconstruction loss using the generator.
Then we use the discriminator to calculate key and value vectors for each real patch and calculate the PCAA loss of fake patches.
To accelerate the training, we only calculate the PCAA loss of 30 uniformly sampled fake patches for one sequence. 
We report additional implementation details in the \cref{app:adv-training}.

\textbf{Balancing MLE and PCAA.}
To mitigate the notorious instability of training a GAN, we follow \citet{taming} and mix both the MLE loss $\mathcal{L}_{\textsf{rec}}$ and our PCAA loss $\mathcal{L}_{\text{PCAA}}$.
We do so by computing
$$
\mathcal{L_{\textsf{stage2}}} = \mathcal{L}_{\textsf{rec}} + \lambda_{\textsf{m}} \cdot \lambda_{\textsf{auto}} \cdot \mathcal{L}_{\text{PCAA}},
$$
where $\lambda_{\textsf{m}}$ is a tunable hyperparameter, and $\lambda_{\textsf{auto}}$ is a value automatically computed as
$\lambda_{\textsf{auto}} = {\nabla_{G_L}[\mathcal{L}_{\textsf{rec}}]}/({\nabla_{G_L} [\mathcal{L}_{\text{PCAA}}] + \delta})$
where $\nabla_{G_L}[\cdot]$ represents the scale of gradients\footnote{The scale is the mean of the element-wise absolute gradients of the output linear projection layer. Because we accumulate the gradients of a batch from mini-batches, we calculate the scale of every mini-batch and then average them.} of the generator w.r.t. its last layer $N$, and $\delta = 1e^{-8}$ is added to avoid division by zero.

\subsection{\model in action}
\label{subsec:tasks}

As previously mentioned, \model is able to tackle generative tasks that are out of the scope of \pixel.
At the same time, it can handle the discriminative tasks used in \citet{pixel}.
Depending on the task, we use \model's learned representations in different ways, as discussed next.

\textbf{Discriminative tasks.} 
Following previous work on downstream discriminative tasks (e.g., \citet{bert}), we add a lightweight prediction head to \model to make it output classification labels or regression scores. 
Because of the causal masking strategy of \model (conditioning on patches from left to right), only the last hidden state $h^{\text{out}}_{\text{eos}}$ associated with the EOS patch can attend over all the patches in the input sequence.
Therefore,
our added head linearly projects $h^{\text{out}}_{\text{eos}}$ to logits then followed by a softmax to output the probability of each class, or used as is for regression tasks.

\textbf{Generative tasks.} 
\model generates $L$ image patches at a time,
containing newly rendered words that complete a given prompt.
We can use \model in generative tasks such as language modeling where, given a text passage rendered as a sequence of image patches, the model generates the next word \citep{lambada}.
Similarly, we can also perform few-shot learning experiments for QA by providing pairs of examples and solutions as rendered prompts \citep{babi}.
We use rendered pipe characters as visual delimiters to separate prompts and generated text (see \cref{fig:gensample} and \cref{sec:gentask}).
Note that common generative NLP benchmarks expect predictions as (sequences of) symbolic tokens.
We discuss next how we can extract text from generated pixel patches, and quantify the readability issue linked to MLE. %

\textbf{Text recognition.} 
We concatenate the generated patches into one image and use OCR software to recognize the text within it.
We found, however, that common OCR tools expect higher resolution images to be accurate, and furthermore, they do not work well with binary images, which leads to incorrect recognition even if they are readable by humans.
To improve the OCR accuracy, we scale the generated image patches by a factor of 3 and place them in the middle of a square white image background.
Also, we combine the results of PaddleOCR\footnote{PaddleOCR is available \href{https://github.com/PaddlePaddle/PaddleOCR}{here}} and Tesseract OCR\footnote{Tesseract OCR is available \href{https://github.com/tesseract-ocr/tesseract}{here}}.
If the recognized word from any of the OCRs matches the target word, we count it as a correct prediction.

\textbf{Readability metric.}
As discussed in \cref{sec:readability},
some of the generated patches can be noisy and unreadable by both humans and OCR tools.
To quantify the quality of the generated text by \model, we define readability as the ratio of generated image patches that contain at least one leading ``legal'' English word, i.e., a word that could be recognized by the OCR software and that appears in a reference vocabulary built
by using 333k most common words from the English Word Frequency dataset\footnote{The vocabulary file could be downloaded \href{https://www.kaggle.com/datasets/rtatman/english-word-frequency}{here}}.
As expected, this metric is correlated to higher prediction accuracy (\cref{tab:geneval}).

\section{Experiments}
\label{sec:experiments}

We aim to answer the following research questions: \textbf{RQ1)} how does \model compare w.r.t., \pixel on discriminative tasks?
\textbf{RQ2)} how does it perform as a generative model?
\textbf{RQ3)} how robust is generated text to orthographic attacks?
and \textbf{RQ4)} 
what does \model attend to in a prompt if no symbolic tokens are provided to it as in classical LLMs?

\subsection{Experimental setup}
\label{subsec:render}

\textbf{Data.}
\model is trained using the same pretraining dataset as \pixel, consisting of Bookcorpus~\citep{bookcorpus} and English Wikipedia (see details in \citet{pixel}).
We follow their preprocessing as described in Appendix \ref{app:datapreprocess}. 

\textbf{Text rendering.}
We use the same PangoCairo render of \pixel,
but using $H=W=8$ for the division in patches.
For better readability under this low resolution, we use the pixel-style font ``Pixeloid Sans''.\footnote{The font file is available at the following \href{https://www.dafont.com/pixeloid-sans.font}{link.}}
In initial experiments, we found this to be performance-neutral compared with Noto font used by \pixel.
For binary images, we render each pixel to a gray-scale value first and then binarize each pixel value by applying a threshold $\theta = 0.5$.
For RGB images, we map each pixel value to the [0, 1] interval.
We follow \pixel's implementation that uses a black patch to represent special EOS tokens,
which we use also for delimiting pairs of sentences.

\textbf{Design choices and ablations.}
Due to computational constraints, we run preliminary experiments to determine 
\textbf{a)} the number $L$ of patches to predict during training, \textbf{b)} whether to use a linear projection layer to map from and to pixel space or perform inference in latent space \citep{ldm}, and \textbf{c)} RGB or binary encoding for images.
We found that $L=2$ in binary (linear) pixel space yields the best results and we adopt it for the rest of the experiments.
\cref{sec:ablations} details this process.

\textbf{Stage 1 pretraining.}
In stage 1, \model is optimized by the AdamW optimizer \cite{adamw} for 1M steps with batch size set to 384.
We linearly warmed up the learning rate to  3e-4 in the first 2000 steps and then annealed it to 3e-6 using a cosine scheduler \cite{cos}.
For discriminative tasks, we pretrained a \model with 85M parameters, which is equivalent to the size of the encoder of the \pixel.
For generative tasks, we used a \model with 113M parameters instead to have a comparable size with GPT-2.
We report details of pretraining and model architecture hyperparameters in \cref{app:pretraindetail}.

\textbf{Stage 2.}
We experimented with different values of 
$\lambda_{\textsf{m}}$. Specifically, we trained the stage 1 model for 200 steps using different $\lambda_{\textsf{m}}$ from 0.1 to 15, selecting the one with best validation performance.

\subsection{RQ1) Discriminative Tasks}
\label{subsec:glue}
\begin{table*}[!t]
\centering
\small
\scalebox{0.95}{
\begin{tabular}{lccccccccccc}
\toprule
\multirow{2}{*}{model} & \multirow{2}{*}{$|\theta|$} & MNLI-m/mm & QQP  & QNLI & SST-2 & COLA & STSB & MRPC & RTE  & WNLI & \multirow{2}{*}{AVG} \\
                       &                             & 392k      & 363k & 108k & 67k   & 8.5k & 5.7k & 3.5k & 2.5k & 635  &                      \\ \midrule
BERT                   & 110M                        & 84.0/84.2 & 87.6 & 91.0 & 92.6  & 60.3 & 88.8 & 90.2 & 69.5 & 51.8 & 80.0                 \\ 
GPT-2                   & 126M                        & 81.0 & 89.4 & 87.7 & 92.5  & 77.0 & 74.9 & 71.5 & 52.0 & 54.9 & 75.6                 \\ \midrule

\pixel                & 86M                         & 78.1/78.9 & 84.5          & \textbf{87.8} & \textbf{89.6}  & 38.4          & 81.1          & \textbf{88.2} & \textbf{60.5} & 53.8          & 74.1                \\ 
\model (stage 1)                 & 85M                         & 78.4/78.6 & 85.6 & 85.7          & 89.0           & \textbf{39.9} & 81.7 & 83.3          & 58.5          & 59.2 & 74.0                 \\ 
\model (stage 2)                 & 85M                         &\textbf{79.7}/\textbf{80.1}  &\textbf{86.3}  &85.7          &89.3            &37.0  &\textbf{82.4}  &82.8           &57.7           & \textbf{60.6} & \textbf{74.2}               \\ 
\bottomrule
\end{tabular}}
\caption[Caption for LOC]{\textbf{\model achieves on-par performance with \pixel on GLUE} with a simpler structure. We report the F1 score for MRPC and QQP, Matthew's correlation for COLA, Spearman's $\rho$ for STSB, and accuracy for other tasks. We report BERT performance here as a reference. Scores of BERT and \pixel are originally reported in \citet{pixel}. GPT-2 scores are reported from Huggingface. The best performance of each task is marked in bold.
}
\label{tab:glue}
\end{table*}

We follow \citet{pixel} and evaluate the language understanding ability of \model using the GLUE benchmark.
GLUE consists of 8 classification tasks and 1 regression task.
For each task, we finetune the pretrained \model with a newly initialized prediction head using rendered data.
The rendering configuration is the same as the pretraining.
In some tasks, an example consists of a pair of sentences. To delimit the two sentences, we inserted a black patch between them.
As discussed in Section \ref{subsec:tasks}, we used the embedding from the last black patch as the input of the task head.

We adopt the same hyperparameters for fine-tuning used by \citet{pixel}. 
Additionally, during training, we use an early-stopping strategy based on the validation set of each corresponding dataset.
For tasks with more than 300k samples, e.g. MNLI and QQP, we set the maximum training step as 8000 and batch size as 256.
We found that \model is robust to hyperparameters on tasks with more data and sensitive to batch size or learning rate for smaller datasets (see Appendix \ref{app:gluedetail}).

As shown in Table \ref{tab:glue}, \model achieves on-par average performance with \pixel (74.0 vs 74.1) with a substantially simpler architecture, and scores very closely to GPT-2 (75.6). 
Note that 
\pixel is a 112M encoder-decoder model but only the 86M encoder is used in this experiment and GPT-2 uses even more parameters (126M). We also highlight the stark improvement of 5.4 points on WNLI compared to \pixel. Considering that WNLI is based on fictional books, the improvement in performance on WNLI might be associated with \model's ability to learn better representations from the pretraining dataset that includes books (i.e., BookCorpus). 
We also note that stage-2 pretraining slightly increases performance for discriminative tasks. 
According to our intuition, the adversarial training improves \model's hidden states representations making them more suitable for discriminative tasks.

\subsection{RQ2) Generative Tasks}
\label{sec:gentask}

\begin{table}[!t]
\centering
\scalebox{.9}{
\begin{tabular}{lccc}
\hline
Model       & $|\theta|$ & LAMBADA  & bAbI  \\ \hline
GPT-2        & 124M       & 17.1     & 26.8 \\ 
$\model_{\text{(stage1)}}$  & 113M    & 5.7 (54.8)     & 11.1 (63.2) \\ 
$\model_{\text{(stage2)}}$  & 113M    & 13.8 (82.2)    & 19.6 (77.0) \\ \hline
\end{tabular}
}
\caption{
\textbf{\model with a 2-stage pretraining is comparable with GPT-2 on short generative tasks.} We report the zero-shot last word prediction accuracy and readability ratio (in brackets) of the LAMBADA test set and the few-shot accuracy of 10K synthesized samples from bAbI. 
}
\label{tab:geneval}
\end{table}

For generative tasks, we render the prompt into an image and insert a white space patch of 3 pixels long to start the generation---intended as a reasonable space to delimit the beginning of a new word. 
Then \model generates the next image patches autoregressively from there. 
We use two datasets for the generative tasks evaluation:
\textbf{LAMBADA}, a benchmark designed to evaluate the text-understanding capability of LLMs where 
models have to predict the last word of a sentence in a given context, and \textbf{bAbI} a QA task that evaluates the model's reading comprehension ability on some given facts.
For bAbI, since \model and GPT-2 are not directly trained on QA data, we show each model 4 examples in the prompt and split the question and answer using the ``\textbar" symbol (see \cref{fig:gensample} and \cref{subsec:tasks}).

In LAMBADA the model needs to capture long-range dependencies in a context that is exposed only to the models during evaluation.
From Table \ref{tab:geneval}, we can observe that after only 200 steps of the second stage of pretraining, we can boost the performance of \model and its overall generation readability (\cref{subsec:tasks}) by 8.1 points.
Although \model cannot beat GPT-2, it is almost in the same ballpark (13.8 versus 17.1), but it is 27\% smaller than GPT-2 and our automatic OCR pipeline might not perfectly recognize some differently rendered words. 
For bAbI, we confirm the benefit of stage 2 pretraining (as we can also generate 77\% of ``readable'' answers), but \model's final performance is a bit further away from GPT-2 (19.6 vs 26.8), hinting that it is a more challenging task than LAMBADA.
Nevertheless, our results on bAbI provide a promising signal that few-shot learning is also possible with pixel-based generative models such as \model.
We complement this experiment by showing more generation examples in \cref{fig:gensample,fig:gensamplemore} and in our analysis of attention in \cref{sec:attention}.

\begin{figure}[!tb]
   \begin{minipage}{0.24\textwidth}
     \centering
     \includegraphics[width=\linewidth]{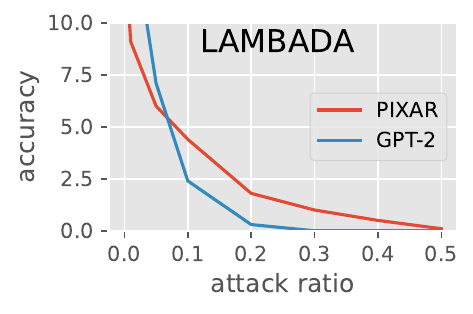}
   \end{minipage}\hfill
   \begin{minipage}{0.24\textwidth}
     \centering
     \includegraphics[width=\linewidth]{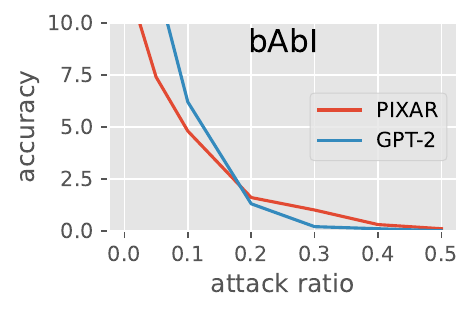}
   \end{minipage}
   \caption{\textbf{\model is more robust than GPT-2 under high visual attack ratios}, especially on LAMBADA. We measured zero-shot accuracy on LAMBADA and few-shot accuracy on bAbI.
   }
   \label{fig:attack}
\end{figure}

\subsection{RQ3) Robustness to orthographic attacks}\label{sec:attacks}

Following \citet{pixel}, we use the procedure used in \citet{zero} to evaluate whether \model can be more robust to ``visual attacks" than GPT-2 when performing generative tasks such as LAMBADA and bAbI. 
First, we manually select a subset of look-alike characters for every English letter from the visually confusable characters of Unicode Technical Standard \#39 set~\footnote{Confusables characters can be found \href{https://util.unicode.org/UnicodeJsps/confusables.jsp}{here}.}. In this way, we make sure that the characters used in the visual attacks can be displayed with the ``Pixeloid Sans'' font.
During the evaluation, we randomly replaced a ratio of letters in a prompt with a random look-alike character and measured the accuracy of the generated word. The results in Figure \ref{fig:attack} demonstrate that although both models suffer from the visual attack, \model showcased higher robustness under higher visual attack ratios. We report details of the performance under visual attack in Table \ref{tab:attack}.

\subsection{RQ4) What is \model looking at?}
\label{sec:attention}
\begin{figure*}[!t]
\centering
\small
\setlength{\tabcolsep}{2pt}
    \begin{tabular}{cccc}
         \rotatebox{90}{
         $1^{\mathsf{st}}$ layer
         }& 
         \includegraphics[width=.465\textwidth]{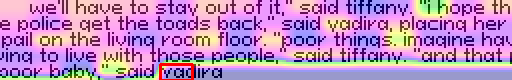}
         &
         \rotatebox{90}{
         $3^{\mathsf{rd}}$ layer
         }
         &
         \includegraphics[width=.465\textwidth]{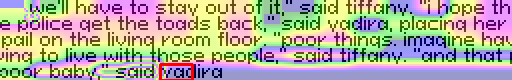}\\
         \rotatebox{90}{
         $5^{\mathsf{th}}$ layer
         }& 
         \includegraphics[width=.465\textwidth]{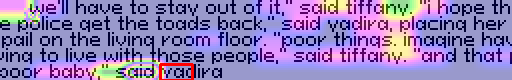}
         &
         \rotatebox{90}{
         $7^{\mathsf{th}}$ layer
         }
         &
         \includegraphics[width=.465\textwidth]{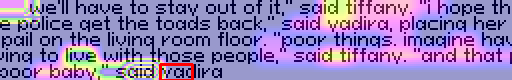}\\
         \rotatebox{90}{
         $9^{\mathsf{th}}$ layer
         }& 
         \includegraphics[width=.465\textwidth]{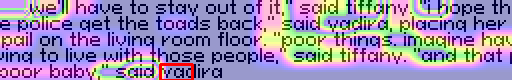}
         &
         \rotatebox{90}{
         $11^{\mathsf{th}}$ layer
         }
         &
         \includegraphics[width=.465\textwidth]{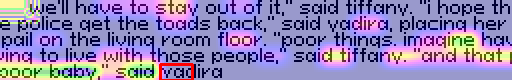}
         \end{tabular}

    \caption{\textbf{\model looks at longer patch sequences in the first layers and then focuses on specific word-like sequences} as shown by the above heatmaps of the attention weights for the first generated patch of ``yadira''. The same pattern is visible in many more examples in \cref{fig:attn layer more}.}
    \label{fig:attn layer}
\end{figure*}

\begin{figure*}[!t]
\centering
\small
\setlength{\tabcolsep}{2pt}
    \begin{tabular}{cccc}
         \rotatebox{90}{
         $1^{\mathsf{st}}$ patch
         }& 
         \includegraphics[width=.465\textwidth]{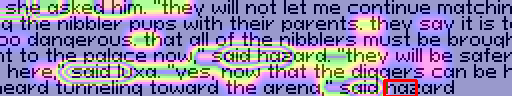}
         &
         \rotatebox{90}{
         $2^{\mathsf{nd}}$ patch
         }
         &
         \includegraphics[width=.465\textwidth]{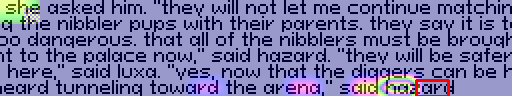}
         \end{tabular}
    \caption{\textbf{\model focus on more context patch sequences when generating the first patch of a word, then narrowing its focus}, as shown by the above heatmaps of the attention weights for two consecutively generated patches of the correctly predicted "hazard". 
    \cref{fig:attn patch more} shows more examples of the same pattern
    .}
    \label{fig:attn patch}
\end{figure*}

\begin{figure*}[!t]
    \centering
    \small
\setlength{\tabcolsep}{2pt}
    \begin{tabular}{ccccc}
    \rotatebox{90}{
         \textsc{correct}
         }
         \rotatebox{90}{
         rope
         }& 
         \includegraphics[width=.465\textwidth]{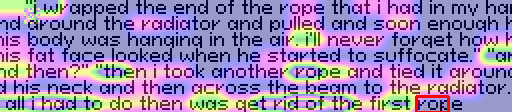}
         &
         \rotatebox{90}{
         Leo
         }
         &
         \includegraphics[width=.465\textwidth]{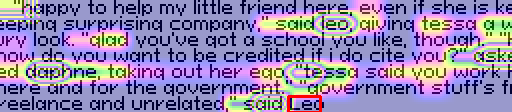}\\
         \rotatebox{90}{
         \textsc{incorrect}
         }
         \rotatebox{90}{
         coffee
         }& 
         \includegraphics[width=.465\textwidth]{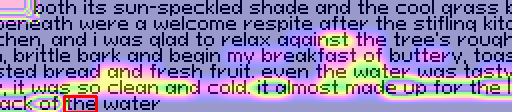}
         &
         \rotatebox{90}{
         bob
         }
         &
         \includegraphics[width=.465\textwidth]{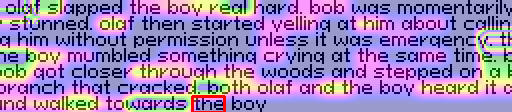}
         \end{tabular}
    \caption{
    \textbf{\model attends to sub-words that can be semantically relevant to generate correct answers} (top), even when the generated answer is wrong (bottom). Correct answers are on the left of each figure.
    \cref{fig:attn right more} contains more examples. %
    }
    \label{fig:attn right wrong}
\end{figure*}

Modeling text purely from perceptual information raises the question of how \model processes such information to generate configurations of pixels that resemble symbolic information (words), some of which might not have been seen during training.
This question is vast (and challenging!) and, in this work, we attempt to answer it by looking at which parts of the rendered pixels \model attends to.
To do so, we take inspiration from the attention map analysis of \citet{pixel}, and inspect \textbf{RQ4.1)} how attention changes through the Transformers layers,
\textbf{RQ4.2)}, what \model is looking at when generating one patch at a time, and
\textbf{RQ4.3)} if there is any correlation with (sub-)words that are semantically meaningful for the generative task at hand.
For all the above questions, we examine the attention weights when generating image patches given prompts from the LAMBADA task.
For each generated image patch, we calculate the average attention weights across all attention heads of all layers (RQ4.1) or just the last layer (RQ4.2,3).
We then plot these averaged weights as heatmaps, which allow us to observe the areas of focus during the generation process.
We provide some examples in \cref{fig:attn layer,fig:attn patch,fig:attn right wrong}, and more in \cref{app:heatmap}.

\textbf{RQ4.1)} We visualize the attention weights of all layers when generating the first image patch (\cref{fig:attn layer}). 
A consistent pattern we observe is that the bottom layers (e.g. first and third layers) tend to focus on many parts of the input prompt at once.
Layers closer to the final one instead attend more intensely to sub-words, filtering out irrelevant information. 
This hints at a possible refinement process in \model's attention mechanism as the information propagates throughout layers and focuses on patches representing potentially meaningful words. 

\textbf{RQ4.2)} 
Similarly, looking at attended pixels before and after generating the first patch of a word (\cref{fig:attn patch}), suggests that \model has two distinct modes when generating a long word spanning multiple patches, and it distributes attention differently depending on the stage of generation.
Specifically, when generating the first patch of a long word, \model tends to focus its attention on previous patches, suggesting a reliance on contextual information to decide which word to generate. 
Then, after it commits to that word, long contexts become less relevant and \model systematically looks more at patches closer to the generated word.
This second mode suggests a tendency of \model to preserve consistency and coherence within the ongoing word generation process.

\textbf{RQ4.3)} 
Finally, we visualize attention when \model generates the correct or incorrect answer. 
We notice that when the answer is present in the prompt, \model correctly attends to it with higher importance.
This is the case for correctly predicted and highlighted ``rope'' and ``Leo'', but also the not predicted but still highlighted ``bob'' in \cref{fig:attn right wrong}.
Even when wrongly predicting ``water'' (the correct answer is ``coffee''), the model attends to ``water'' and other properties related to it in the prompt.
More examples in \cref{fig:attn right more} showcase similar patterns.
While not easy to quantify, this hints at the promising ability of \model to be able to extract, perhaps not correct but still semantically meaningful, symbolic information from perceptual one.

\section{Conclusion}
\label{sec:conclusion}

We introduced \model, an autoregressive pixel-based LLM that inherits the benefits of the first purely pixel-based LLM \pixel but, differently from it, can support generative language tasks.
We showcased the capabilities of \model on discriminative tasks such as GLUE where it achieves comparable (or superior in some cases) performance w.r.t. \pixel, and on generative free-text QA tasks, such as LAMBADA and bAbI, where it gets closer to GPT-2 while being more robust to orthographic attacks.
While these results are encouraging, \model performance is still behind more sophisticated LLMs that operate on symbolic tokens.
This is expected, as reading and generating text as pure pixels is a much more challenging task, and we expect to improve on \model's architecture and performance in future work, e.g., by training larger variants of \model, on more text data rendered as images, and with more language variants.
Furthermore, we highlighted how an adversarial loss can greatly improve the quality and accuracy of generated text. 
Overall, our work shed light on the possibility of treating texts as image-domain data paving the way towards more expressive language models that can efficiently generalise across languages and cultures \citep{liu-etal-2021-visually}.

\bibliography{custom}

\newpage
\newpage
\appendix

\section{Limitations \& Risks}
\label{subsec:limitations}

In this paper, we train and evaluate \model on English language data only to compare with other well-known autoregressive language models like GPT-2. \citet{salesky2023multilingual} showed promising results for encoder-only models applied to multilingual data. Due to the simplicity of \model design, we foresee that it can be pretrained using multilingual data following previous work. 
In this paper, we explore the training regime for autoregressive pixel-based language models and find that a simple maximum likelihood regime is suboptimal for these models. Although the short stage 2 adversarial pretraining improved the generation readability and accuracy, \model still has limitations in generating long sentences due to the readability issue.
The notorious instability of adversarial training also makes the result of stage 2 pretraining hard to optimise even with the help of automatic GAN ratio balancing. Therefore, we believe that future work should explore more stable pertaining methods as an important research direction.
Because experiments with \model cover only the English language, we acknowledge that there might be challenges to be solved when tackling different writing systems. 

\section{More related works}
\label{app:byte-level}

\subsection{Generative Language Models}
Language models learn to predict probability distributions over a sequence of input elements which can have different granularity including words, tokens, characters, or pixels, which is the approach we follow in this work. This task is generally referred to as next token prediction and it represents a crucial problem in natural language processing \citep{shannon1948mathematical}. The advent of transformers brought about a huge improvement in this field, especially in parallel training and capturing long-range dependencies \citep{transformers}. Surprisingly, the family of generative pre-trained transformers \citep{gpt2,gpt3, bloom, llama, llama2, felcon} demonstrated their strong ability to understand and generate natural language. 

\subsection{Transformer-based Image Generative Models}
The task of image generation can be formulated as pixel-based autoregressive modeling. In the literature, this task was attempted using different architectures such as CNNs \citep{salimans2017pixelcnn}, RNNs \citep{oord2016pixel} and Transformers \citep{chen2020generative,menick2018generating}. However, the method of generating only 1 pixel at each step is limited to low-resolution images but not high-resolution images due to long sequence lengths and poor fidelity. 
Also, the sequential generative modeling and teacher-forcing training of generative transformers are not suitable for flattened image vectors since they learn to predict the probability distribution over a fixed finite vocabulary instead of image patches. 

To mitigate this issue, a two-stage approach is proposed to learn a Vector-Quantized Variational Autoencoder(VQ-VAE) \citep{vqvae} to map continuous pixel values into a sequence of discrete tokens and then use a transformer decoder to model the distribution of the latent image tokens. This two-stage image tokenization approach not only contributes to generative modeling for high-fidelity images \citep{chang2023musevqvae,chang2022maskgitvqvae,ramesh2022hierarchical, tschannen2023givt} but also shed the way for interleaved multimodal generation by simply concatenating the vocabularies for both text and images \citep{aghajanyan2022cm3,aghajanyan2023scaling}. Learning from the discrete latent space of VQ-VAE becomes a popular way for models to improve image generation quality and enable language-inspired self-supervised learning \citep{bao2022beit, li2023mage}.

\subsection{Byte-level tokenizers}
The idea of using character-level visual features is not limited to pixel-based models. For instance, MacBERT \citep{macbert} and \citet{enchar} also used character-level encodings. However, they still relied on traditional ID-based embeddings and thus suffered from the vocabulary bottleneck problem.

Alternatively, byte-level tokenization is also capable of encoding any code sequence. 
\citet{mt5} trained mT5 on UTF-8 byte sequence to adapt to different languages. 
Perceiver \citep{perceiver} proposed an iterative transformer structure which attended information directly through raw byte arrays. The increased sequence length poses challenges for Transformer-based architectures. For this reason, MEGABYTE~\cite{yu2023megabyte} proposes a hierarchical architecture where a low-level patch decoder is conditioned on a global-level learned patch representations. Another line of research uses recent state space models to extrapolate over extremely long sequence lengths. This is the case for Mamba byte~\cite{wang2024mambabyte} which pretrains a Mamba model~\cite{gu2023mamba} using byte representations.

\section{Data Preprocessing}
\label{app:datapreprocess}

\begin{table}[!h]
\centering
\scalebox{.9}{
\begin{tabular}{lrr}
\toprule
{Dataset} & \multicolumn{1}{c}{\#samples}  & avg. \#characters \\ \midrule
English Wikipedia            & 6,458,670                     & 3,028            \\
Bookcorpus                   & 17,868                        & 370,756          \\
processed                    & 26,759,562                    & 987            \\
\bottomrule
\end{tabular}
}
\caption{\label{tab:pretraindata} Pretraining dataset statistics. 
}
\end{table}

\begin{algorithm*}[hbt!]
    \caption{Text Segmentation}
    \label{alg:sentsplit}
\begin{algorithmic}
    \STATE {\bfseries Input:} text $s$, maximum length $l_{max}$, minimum length $l_{min}$
    
    \STATE Initialize $sentList = SentTokenize(s)$, $sampleList = EmptyList()$, $sample = EmptyString()$
    \FOR{$i=1$ {\bfseries to} $||sentList||$}
        \IF{$len(sample) + len(sentList_i) > l_{max}$}
            \IF {$len(sample) \geq l_{min}$}
                \STATE $Append(sampleList, sample)$
            \ENDIF
            \STATE $sample = sentList_i$
        \ELSE
            \STATE $sample = Concat(sample, sentList_i)$
        \ENDIF
    \ENDFOR
    \IF{$len(sample) \geq l_{min}$}
        \STATE $Append(sampleList, sample)$
    \ENDIF
    \STATE {\bfseries Output:} $sampleList$
\end{algorithmic}
\end{algorithm*}

Because the length of Wikipedia articles and books is usually longer than \model's context window.
We segment each article or book into several samples using the algorithm \ref{alg:sentsplit}.
In particular, we use the ``PunktSentenceTokenizer" from the Natural Language Toolkit (NLTK) \cite{nltk} to segment each article or book into sentences.
Then, we concatenate short sentences to training samples within a character limit.
Finally, we filter out samples with less than 100 characters. Table \ref{tab:pretraindata} demonstrates the length statistics of processed data.

During the pretraining, due to computational constraints, we fix the window size of \model as 360 patches.
Thus, the model trained with a longer patch length can process more texts in each forward pass.
We manually selected 1180 as the character limit for patch length 2. 

\section{Computational budget}

The \model's pretraining stage was completed in $\sim$4 days using 16 NVIDIA V100 GPUs on an HPC computing cluster by extending the codebase originally released by \citet{pixel}. Finetuning experiments are shorter and use the same computing infrastructure.

\section{Hyperparameter Ablations}\label{sec:ablations}
In our preliminary experiments, we explored several design choices for a \model with 113M parameters, including different patch lengths, CNN-based projection layers, and the different formats of data.
Because of our limited computational budget, we experimented with a selected number of variants that could allow us to make informed decisions about our model design.
For each setting, we conducted 0.1M stage 1 pretraining steps and evaluated their performance on the GLUE benchmark \citep{glue}.

\textbf{Patch Length Selection}
\label{subsec:patchlen}

In this ablation, the height of image patches is fixed at 8 pixels and we render all texts as a long image with 8 pixels high.
Thus, the length of each patch defines how many characters or words a patch can hold and also determines the complexity of predicting the next patch.

We pretrained models on images with patch length [2, 5] for 0.1M stage 1 steps.
The results in Table \ref{tab:ablation} show that in all combinations, patch length 2 achieved the highest average score.
Thus, we used patch length 2 as our reference value for all the experiments.

\textbf{Linear Projection vs CNN}
\label{subsec:enmedding}

We experimented with two different strategies to convert image patches to vectors: 1) linear projection of the patches; and 2) a CNN auto-encoder as the input and output layers \cite{ldm}.
We first trained a lightweight CNN auto-encoder with a bottleneck layer of dimension 8 on the pretraining data.
Then we take the encoder part as the model input layer, and the decoder as the output layer.
During the pretraining, the \model backbone is trained in the latent space of the auto-encoder \cite{ldm}. 
We freeze parameters from the auto-encoder and train \model to minimize the MSE of the next latent vector rather than the next image patch. However, as results shown in Table \ref{tab:ablation}, the extra auto-encoder did not bring significant improvement compared with the simple linear projection layer. Therefore, we opted for the simpler linear projection instead.

\textbf{RGB vs binary images}
\label{subsec:convsdis}

As detailed in Section \ref{sec:method}, we explored training with both continuous RGB images and binary images.
When we use linear projection layers, we flatten the values of each channel of the RGB image into a vector as the input and calculate the MSE between the predicted vector and the ground truth vector as the loss.
When using the CNN auto-encoder as input and output layers, RGB image patches are converted to latent vectors with CNN layers with corresponding layers.

The results in Table \ref{tab:ablation} indicate that models trained with continuous RGB images have $~\sim$1.4 points lower than their binary counterparts. 
We therefore used binary rendering in further pretraining experiments.

\section{Licensing}

As mentioned before, the starting point for our work was the \pixel codebase released on \href{https://github.com/xplip/pixel/}{GitHub} under Apache 2.0 license. We also used the same pretraining datasets that was released on Huggingface.
Upon acceptance, we will release our codebase on GitHub following the same approach and license.

\section{Pretraining Details \& Model Configurations}
\label{app:pretraindetail}

\begin{table*}[]
\centering
\begin{tabular}{lc|lc|lc}
\hline
\multicolumn{2}{c|}{Pretrain Hyperparameters} & \multicolumn{2}{c|}{Render Configuration} & \multicolumn{2}{c}{Model Structure} \\ \hline
peak lr              & 3e-4                   & patch length        & 2                   & \#layers               & 12         \\
lr scheduler         & CosineAnnealing        & \#patches           & 360                 & \#attention heads      & 12         \\
min. lr              & 3e-5                   & max \#char.         & 1180                & hidden size            & 768        \\
optimizer            & AdamW                  & min. \#char.        & 100                 & activation             & SwiGLU     \\
$\beta_1$            & 0.9                    & patch height        & 8                   & intermediate size      & 2048 / 3072      \\
$\beta_2$            & 0.95                   & render DPI          & 80                  & \#parameters           & 85M / 113M       \\
weight decay         & 0.1                    & font size           & 8                   &                        &            \\
steps                & 1M                     & font                & PixeloidSans        &                        &            \\
warm up              & 2000                   & binary              & true                &                        &            \\
batch size           & 384                    & Temperature (T)                   & 1                    &                        &            \\
precision            & fp16 \& fp32           & Threshold ($\theta$)              & 0.5                    &                        &            \\
random seed          & 42                     &                     &                     &                        &            \\ \hline
\end{tabular}
\caption{\model pretrain configuration.}
\label{tab:pretraindetail}
\end{table*}

Table \ref{tab:pretraindetail} demonstrates the stage 1 pretraining hyperparameters, render configuration, and model architecture details of \model.

\section{GLUE Finetuning Details}
\label{app:gluedetail}

\begin{table*}[!t]
\centering
\small
\begin{tabular}{rcccccccc}
\scshape
\\ \toprule
Embedding    & \multicolumn{4}{c}{CNNAutoencoder}                                               & \multicolumn{4}{c}{Linear projection}                                                                   \\ \cmidrule(lr){2-5} \cmidrule(lr){6-9}
Image type   & 
Binary & Binary & RGB
& 
RGB & RGB
& 
Binary & Binary & RGB \\ \cmidrule(lr){2-5} \cmidrule(lr){6-9}
Patch length $L$ & 
5 & 2& 2& 5& 2 &2& 
5 & 5             \\ \midrule
MNLI-m/mm    & 72.5/74.1                 & 75.8/76.2                 & 75.8/76.4         &72.4/73.4        & 74.9/76.3                 & \textbf{76.4/77.6}        & 72.5/74.0                 & 72.0/73.1     \\
QQP          & 81.2                      & \textbf{87.8}             & 83.4              &81.8             & 83.5                      & 84.3                      & 82.1                      & 82.3          \\
QNLI         & 82.4                      & \textbf{84.3}             & 84.1              &81.7             & 83.5                      & 84.2                      & 82.8                      & 81.6          \\
SST-2        & 83.3                      & 86.2                      & 86.2              &84.2             & 86.6                      & \textbf{88.0}             & 82.6                      & 83.3          \\
COLA         & 15.2                      & \textbf{30.5}             & 26.6              &13.2             & 30.4                      & 27.7                      & 16.7                      & 10.8          \\
STSB         & 69.2                      & 74.2                      & 74.7              &68.9             & 73.5                      & \textbf{81.2}             & 71.5                      & 75.2          \\
MRPC         & 81.2                      & 83.4                      & 82.9              &82.4             & 81.8                      & 82.5                      & 82.7                      & \textbf{84.2}          \\
RTE          & 57.0                      & 59.6                      & 56.3              &61.4             & 54.9                      & 58.5                      & \textbf{61.7}             & 58.1          \\
WNLI         & \textbf{57.7}             & 56.3                      & \textbf{57.7}     &56.3             & 56.3                      & 56.3                      & 56.3                      & 56.3 \\ \midrule
AVG          & 67.4                      & 71.4                      & 70.4              &67.6             & 70.2                      & \textbf{71.6}             & 68.3                      & 67.7          \\ \bottomrule
\end{tabular}
\caption{\textbf{\model trained with binary data, linear projection, and patch length 2 achieved the best GLUE performance} in our ablation tests where we compare variants pretrained and finetuned (only stage 1) with 0.1M steps and 113M parameters with identical hyperparameter selected heuristically as described in Section \ref{subsec:glue} 
We therefore select this best setting and use it for the full 1M step pretraining.
}
\label{tab:ablation}
\end{table*}

Table \ref{tab:gluehyperparam} displays the hyperparameter details of GLUE finetuning.

\begin{table*}[]
\centering
\begin{tabular}{c|ccccccccc}
\hline
 \model (stage 1)                & MNLI & QQP  & QNLI & SST-2 & COLA & STSB & MRPC & RTE  & WNLI           \\ \hline
lr               & 3e-5 & 3e-5 & 3e-5 & 3e-5  & 3e-5 & 3e-5 & 6e-5 & 3e-5 & 3e-5           \\
Optimizer        & \multicolumn{9}{c}{AdamW}                                               \\
$\beta_1$        & \multicolumn{9}{c}{0.9}                                                 \\
$\beta_2$        & \multicolumn{9}{c}{0.95}                                                \\
weight decay     & 0.1  & 0.1  & 0.1  & 0.01  & 0.01 & 0.01 & 0.01 & 0.01 & 0.01           \\
warmup           & \multicolumn{9}{c}{Linear Warmup}                                       \\
warmup steps     & 1000 & 1000 & 500  & 200   & 50   & 100  & 20   & 50   & 2              \\
max steps        & 8000 & 8000 & 4000 & 2000  & 500  & 2000 & 500  & 500  & 20             \\
batch size       & 256  & 256  & 256  & 256   & 256  & 32   & 64   & 32   & 128            \\
evaluation freq. & 500  & 500  & 200  & 200   & 100  & 100  & 50   & 50   & $\sim$ 1 epoch \\
random seed      & \multicolumn{9}{c}{42}                                                  \\ \hline
 \model (stage 2)                & MNLI & QQP  & QNLI & SST-2 & COLA & STSB & MRPC & RTE  & WNLI           \\ \hline
lr               & 3e-5 & 3e-5 & 3e-5 & 3e-5  & 3e-5 & 3e-5 & 3e-5 & 6e-5 & 3e-5           \\
Optimizer        & \multicolumn{9}{c}{AdamW}                                               \\
$\beta_1$        & \multicolumn{9}{c}{0.9}                                                 \\
$\beta_2$        & \multicolumn{9}{c}{0.95}                                                \\
weight decay     & 0.1  & 0.1  & 0.1  & 0.01  & 0.01 & 0.01 & 0.01 & 0.01 & 0.01           \\
warmup           & \multicolumn{9}{c}{Linear Warmup}                                       \\
warmup steps     & 1000 & 1000 & 500  & 200   & 50   & 100  & 20   & 50   & 2              \\
max steps        & 8000 & 8000 & 4000 & 2000  & 500  & 2000 & 500  & 500  & 20             \\
batch size       & 256  & 256  & 256  & 256   & 256  & 128   & 128   & 128   & 128            \\
evaluation freq. & 500  & 500  & 200  & 200   & 100  & 100  & 50   & 50   & $\sim$ 1 epoch \\
random seed      & \multicolumn{9}{c}{42} \\
\hline
\end{tabular}
\caption{Hyperparameters we used to finetune \model on the GLUE benchmark.}
\label{tab:gluehyperparam}
\end{table*}

\section{Generation Samples}
\label{app:generationsamples}

\begin{figure*}[!t]
\centering
\includegraphics[width=\textwidth]{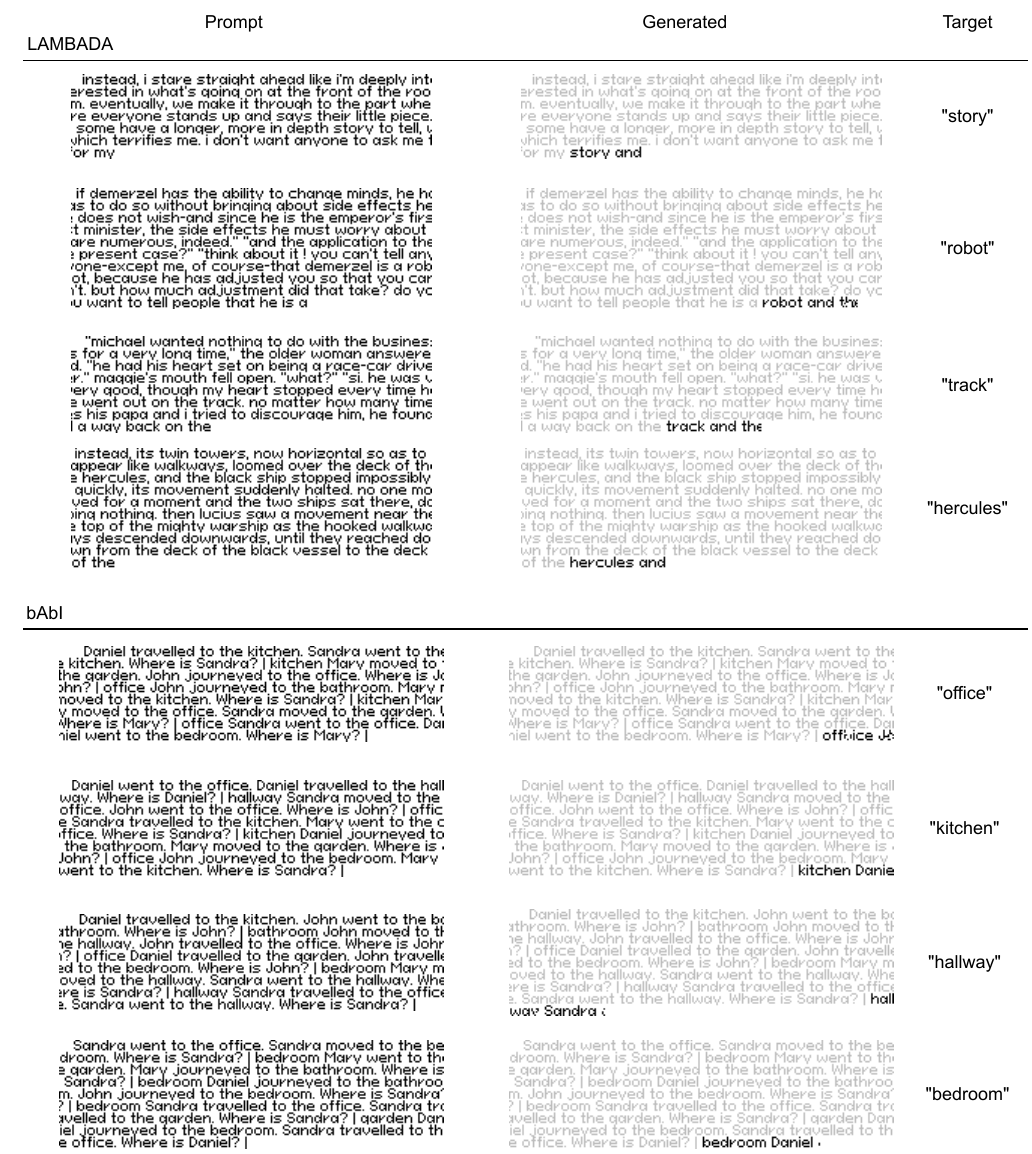}
\caption{\model generation samples from LAMBADA and bAbI dataset. Images are folded into rectangles for better readability.}
\label{fig:gensamplemore}
\end{figure*}

Figure \ref{fig:gensamplemore} shows more generation samples from LAMBADA and bAbI datasets. Images are visualized using a 0.5 threshold and reshaped to a more square size for a better representation.

\section{Attention Heatmap Samples}
\label{app:heatmap}

\begin{figure*}[!t]
\centering
\small
\setlength{\tabcolsep}{2pt}
    \begin{tabular}{cccc}
         \rotatebox{90}{
         $1^{\mathsf{st}}$ patch
         }& 
         \includegraphics[width=.465\textwidth]{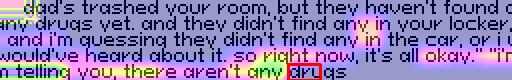}
         &
         \rotatebox{90}{
         $2^{\mathsf{nd}}$ patch
         }
         &
         \includegraphics[width=.465\textwidth]{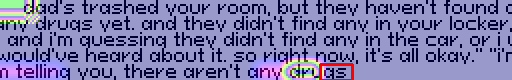}\\
         & \multicolumn{3}{c}{Target: drug}\\
         
         \rotatebox{90}{
         $1^{\mathsf{st}}$ patch
         }& 
         \includegraphics[width=.465\textwidth]{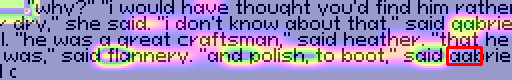}
         &
         \rotatebox{90}{
         $2^{\mathsf{nd}}$ patch
         }
         &
         \includegraphics[width=.465\textwidth]{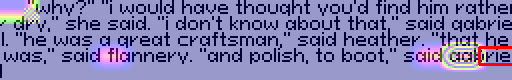}\\
         & \multicolumn{3}{c}{Target: gabriel}\\

         \rotatebox{90}{
         \centering
         $1^{\mathsf{st}}$ patch
         }& 
         \includegraphics[width=.465\textwidth]{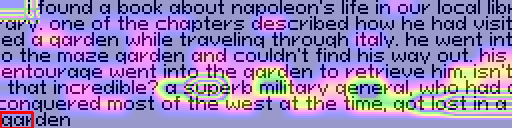}
         &
         \rotatebox{90}{
         $2^{\mathsf{nd}}$ patch
         }
         &
         \includegraphics[width=.465\textwidth]{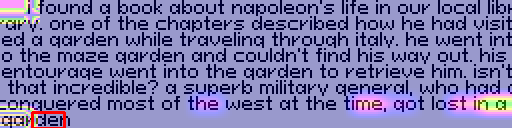}\\
         & \multicolumn{3}{c}{Target: garden}\\

         \rotatebox{90}{
         $1^{\mathsf{st}}$ patch
         }& 
         \includegraphics[width=.465\textwidth]{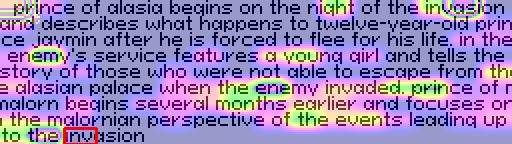}
         &
         \rotatebox{90}{
         $2^{\mathsf{nd}}$ patch
         }
         &
         \includegraphics[width=.465\textwidth]{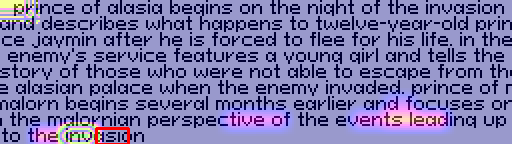}\\
         & \multicolumn{3}{c}{Target: invasion}\\

         \rotatebox{90}{
         $1^{\mathsf{st}}$ patch
         }& 
         \includegraphics[width=.465\textwidth]{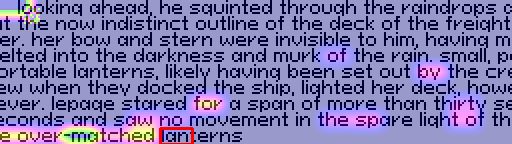}
         &
         \rotatebox{90}{
         $2^{\mathsf{nd}}$ patch
         }
         &
         \includegraphics[width=.465\textwidth]{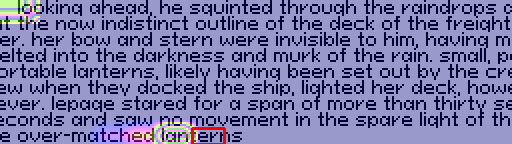}\\
         & \multicolumn{3}{c}{Target: lanterns}\\

         \rotatebox{90}{
         $1^{\mathsf{st}}$ patch
         }& 
         \includegraphics[width=.465\textwidth]{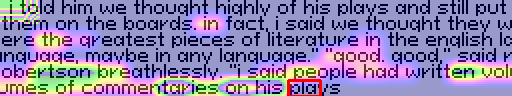}
         &
         \rotatebox{90}{
         $2^{\mathsf{nd}}$ patch
         }
         &
         \includegraphics[width=.465\textwidth]{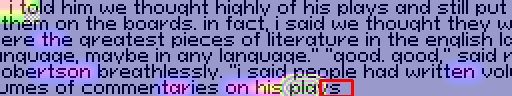}\\
         & \multicolumn{3}{c}{Target: plays}\\

         \rotatebox{90}{
         $1^{\mathsf{st}}$ patch
         }& 
         \includegraphics[width=.465\textwidth]{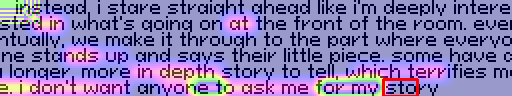}
         &
         \rotatebox{90}{
         $2^{\mathsf{nd}}$ patch
         }
         &
         \includegraphics[width=.465\textwidth]{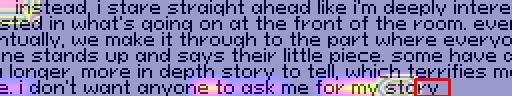}\\
         & \multicolumn{3}{c}{Target: story}\\

         \rotatebox{90}{
         $1^{\mathsf{st}}$ patch
         }& 
         \includegraphics[width=.465\textwidth]{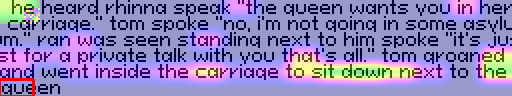}
         &
         \rotatebox{90}{
         $2^{\mathsf{nd}}$ patch
         }
         &
         \includegraphics[width=.465\textwidth]{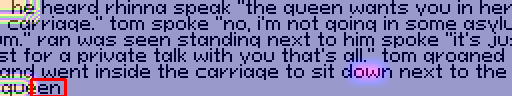}\\
         & \multicolumn{3}{c}{Target: queen}\\
    \end{tabular}
    \caption{Heatmaps of the attention weights of consecutively generated image patches. The attention weights are collected from the last transformer layer of \model. The corresponding image patches of the sub-figure are mark by the red rectangles.}
    \label{fig:attn patch more}
\end{figure*}

\begin{figure*}[!t]
    \centering
    \small
\setlength{\tabcolsep}{2pt}
    \begin{tabular}{ccccc}

    & \multicolumn{3}{c}{\textsc{correct}}\\
    \rotatebox{90}{
         }
         \rotatebox{90}{
         ares
         }& 
         \includegraphics[width=.465\textwidth]{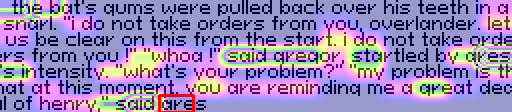}
         &
         \rotatebox{90}{
         border
         }
         &
         \includegraphics[width=.465\textwidth]{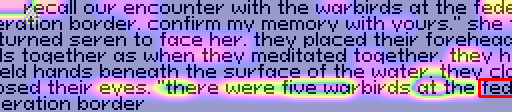}\\
    \rotatebox{90}{
         }
         \rotatebox{90}{
         duke
         }& 
         \includegraphics[width=.465\textwidth]{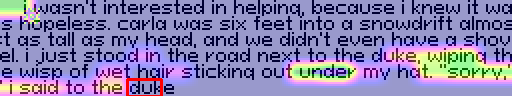}
         &
         \rotatebox{90}{
         gabriel
         }
         &
         \includegraphics[width=.465\textwidth]{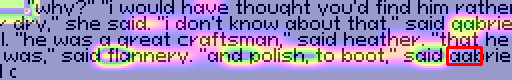}\\
    \rotatebox{90}{
         }
         \rotatebox{90}{
         hercules
         }& 
         \includegraphics[width=.465\textwidth]{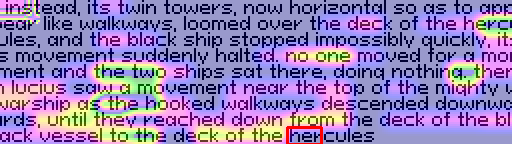}
         &
         \rotatebox{90}{
         greg
         }
         &
         \includegraphics[width=.465\textwidth]{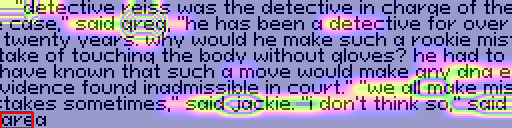}\\
    \rotatebox{90}{
         }
         \rotatebox{90}{
         guy
         }& 
         \includegraphics[width=.465\textwidth]{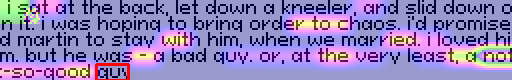}
         &
         \rotatebox{90}{
         list
         }
         &
         \includegraphics[width=.465\textwidth]{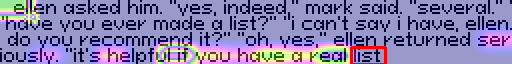}\\
    \rotatebox{90}{
         }
         \rotatebox{90}{
         mareath
         }& 
         \includegraphics[width=.465\textwidth]{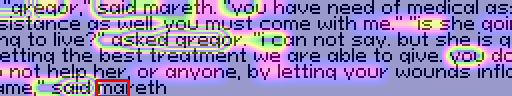}
         &
         \rotatebox{90}{
         mum
         }
         &
         \includegraphics[width=.465\textwidth]{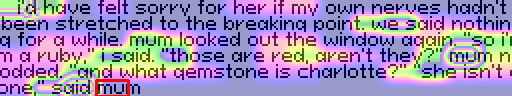}\\
         
        & \multicolumn{3}{c}{\textsc{incorrect}}\\
    \rotatebox{90}{
         }
         \rotatebox{90}{
         admiral
         }& 
         \includegraphics[width=.465\textwidth]{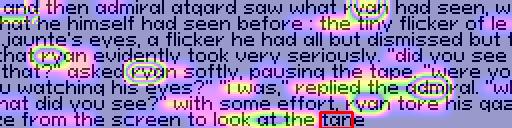}
         &
         \rotatebox{90}{
         daniel
         }
         &
         \includegraphics[width=.465\textwidth]{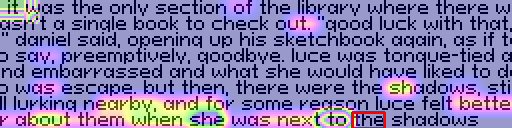}\\
    \rotatebox{90}{
         }
         \rotatebox{90}{
         camera
         }& 
         \includegraphics[width=.465\textwidth]{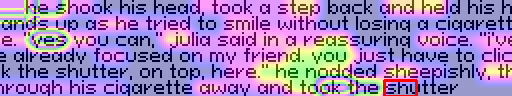}
         &
         \rotatebox{90}{
         gabriel
         }
         &
         \includegraphics[width=.465\textwidth]{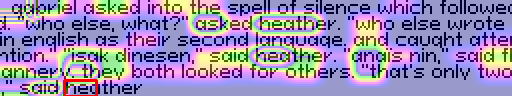}\\
    \rotatebox{90}{
         }
         \rotatebox{90}{
         babies
         }& 
         \includegraphics[width=.465\textwidth]{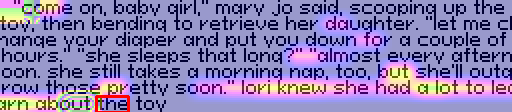}
         &
         \rotatebox{90}{
         moscow
         }
         &
         \includegraphics[width=.465\textwidth]{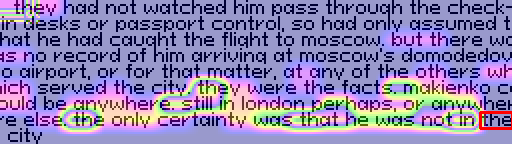}\\
    \rotatebox{90}{
         }
         \rotatebox{90}{
         carlos
         }& 
         \includegraphics[width=.465\textwidth]{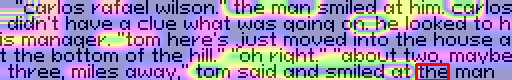}
         &
         \rotatebox{90}{
         sled
         }
         &
         \includegraphics[width=.465\textwidth]{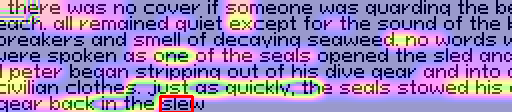}\\
    \end{tabular}
    \caption{Heatmaps of the attention weights of \model in correct and incorrect generations. The attention weights are collected from the last transformer layer of the first generated patches, as marked by the red rectangles.}
    \label{fig:attn right more}
\end{figure*}

\begin{figure*}[!t]
\centering
\small
\setlength{\tabcolsep}{2pt}
    \begin{tabular}{cccc}
         \rotatebox{90}{
         $1^{\mathsf{st}}$ layer
         }& 
         \includegraphics[width=.465\textwidth]{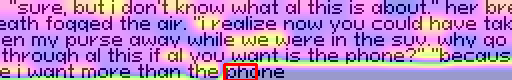}
         &
         \rotatebox{90}{
         $3^{\mathsf{rd}}$ layer
         }
         &
         \includegraphics[width=.465\textwidth]{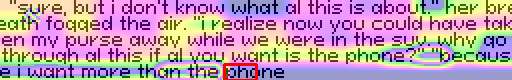}\\
         \rotatebox{90}{
         $5^{\mathsf{th}}$ layer
         }& 
         \includegraphics[width=.465\textwidth]{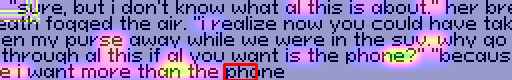}
         &
         \rotatebox{90}{
         $7^{\mathsf{th}}$ layer
         }
         &
         \includegraphics[width=.465\textwidth]{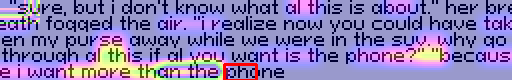}\\
         \rotatebox{90}{
         $9^{\mathsf{th}}$ layer
         }& 
         \includegraphics[width=.465\textwidth]{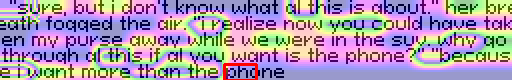}
         &
         \rotatebox{90}{
         $11^{\mathsf{th}}$ layer
         }
         &
         \includegraphics[width=.465\textwidth]{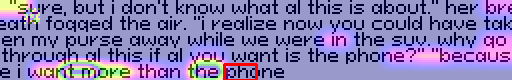}\\
         & \multicolumn{3}{c}{Target: phone}\\
         
         \rotatebox{90}{
         $1^{\mathsf{st}}$ layer
         }& 
         \includegraphics[width=.465\textwidth]{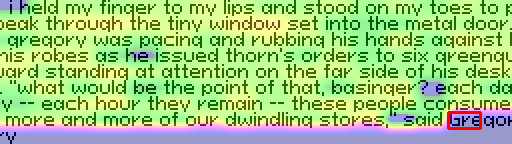}
         &
         \rotatebox{90}{
         $3^{\mathsf{rd}}$ layer
         }
         &
         \includegraphics[width=.465\textwidth]{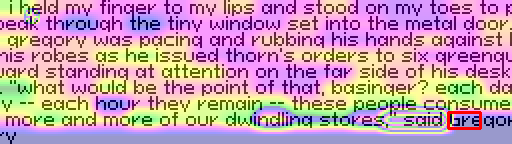}\\
         \rotatebox{90}{
         $5^{\mathsf{th}}$ layer
         }& 
         \includegraphics[width=.465\textwidth]{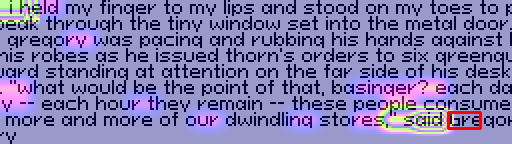}
         &
         \rotatebox{90}{
         $7^{\mathsf{th}}$ layer
         }
         &
         \includegraphics[width=.465\textwidth]{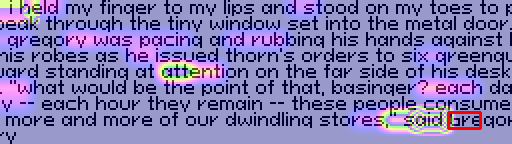}\\
         \rotatebox{90}{
         $9^{\mathsf{th}}$ layer
         }& 
         \includegraphics[width=.465\textwidth]{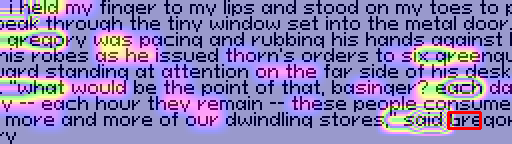}
         &
         \rotatebox{90}{
         $11^{\mathsf{th}}$ layer
         }
         &
         \includegraphics[width=.465\textwidth]{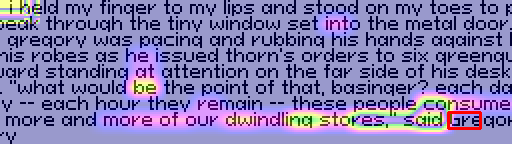}\\
         & \multicolumn{3}{c}{Target: geogory}
    \end{tabular}
    \caption{Heatmaps of the attention weights from the selected transformer layers of first generated image patches, as marked by the red rectangles. }
    \label{fig:attn layer more}
\end{figure*}

Figure \ref{fig:attn patch more}, Figure \ref{fig:attn right more} and Figure \ref{fig:attn layer more} show more attention heatmaps from LAMBADA datasets. The first generated image patches are marked by red rectangles. Samples are drawn using the attention weights from the last transformer layer of the first generated image patch to the given prompt.

\section{Adversarial Training}
\label{app:adv-training}

\begin{figure}[th!]
\centering
\includegraphics[width=0.5\textwidth]{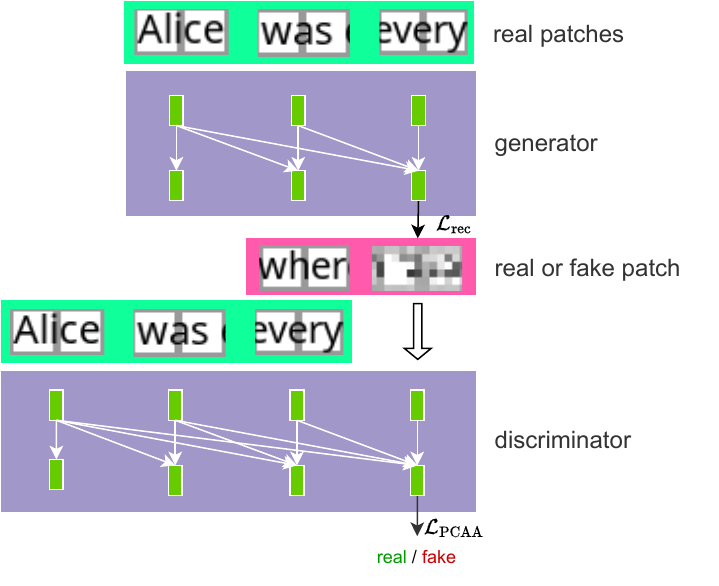}
\caption{\textbf{We copied the \model from stage 1 pretraining as both the generator and the discriminator} for the adversarial training in stage 2. The generator is optimized by both reconstruction loss $\mathcal{L}_{\text{rec}}$ and adversarial loss $\mathcal{L}_{\text{PCAA}}$ to improve the generation readability.
}
\label{fig:stage2}
\end{figure}

\model is trained using a second stage of pretraining using an adversarial training regime that resembles the one used in GANs~\cite{gan}. \cref{fig:stage2} exemplifies how \model is used for the stage 2 pretraining. Each adversarial step can be divided into 3 stages:

\textbf{1) Calculate the Reconstruction Loss}

we fist sample a batch of sequences $\im$ from the pretraining dataset and generate the fake sequences $\tilde{\im}$ using the generator.
Then we calculate the reconstruction loss $\mathcal{L}_{\textsf{rec}}$ as well as generator's gradients w.r.t. $\mathcal{L}_{\textsf{rec}}$.
Because we need to reuse the computational graph later for the PCAA loss, we thus set ``retain\_graph = True" while doing backward propagation using Pytorch.
We record the current gradients of the last generator layer as $\boldsymbol{g}_1$ and calculate the scale of it $\nabla_{G_L}[\mathcal{L}_{\text{rec}}] = |\boldsymbol{g}_1|$.

\textbf{2) Calculate the balanced PCAA and update the generator}

First, we calculate key and value vectors for every real patch using the discriminator and cache them for the PCAA calculation.
To accelerate the calculation, we randomly sample 30 fake patches for each generated sequence and calculate the PCAA loss with the cached key and value vectors using the discriminator.
Then we multiply the PCAA loss with $\lambda_\text{m}$ and $\lambda_{\text{auto}}$ which is calculated from the previous step, and backpropagate all the way to the generator.
Because Pytorch accumulates gradients of multiple backwards steps, to attain the gradients of the last generator layer w.r.t. the PCAA loss, we subtract the previously recorded gradients $\boldsymbol{g}_1$ and calculate the scale $\nabla_{G_L}[\mathcal{L}_{\text{PCAA}}] = |\boldsymbol{g}_2 - \boldsymbol{g}_1| / \lambda_{\text{m}} / \lambda_{\text{auto}}$.
Because the PCAA is balanced, we re-scale it to get the true gradients scale.
The scale is recorded for the next batch.
During the first step, because we initialize $\lambda_{\text{auto}}$ as 1.
Now gradients w.r.t. PCAA and reconstruction loss are ready and we can update the parameters of the generator using the specified optimizer.

\textbf{3) Update the discriminator}

Finally, we calculate the classification loss using the cross entropy loss of the discriminator and update its parameters.
The procedure is similar to the PCAA calculation, we calculate the loss on real patches and cache key and value vectors, then calculate the loss on 30 randomly sampled fake patches for each sequence.

\section{Performance under Visual Attack}

Table \ref{tab:attack} records the accuracy details of GPT-2 and LAMBADA under different visual attack ratios. 
While evaluating the accuracy, we randomly select a ratio of letters from the prompt and replace them with random look-alike symbols. 
If the first word the model predicts matches the target word regardless of its casing, we count it as a correct prediction.

\begin{table}[]
\begin{tabular}{ccccc}
\hline
attack & \multicolumn{2}{c}{LAMBADA} & \multicolumn{2}{c}{bAbI} \\ \cline{2-5} 
ratio  & GPT-2        & PIXAR        & GPT-2       & PIXAR      \\ \hline
0.0    & 17.1         & 13.8         & 26.8        & 19.6       \\
0.01   & 15.0         & 9.1          & 21.6        & 11.4       \\
0.05   & 7.1          & 6.0          & 12.1        & 7.4        \\
0.1    & 2.4          & 4.4          & 6.2         & 4.8        \\
0.2    & 0.3          & 1.8          & 1.3         & 1.6        \\
0.3    & 0.0          & 1.0          & 0.2         & 1.0        \\
0.4    & 0.0          & 0.5          & 0.1         & 0.3        \\
0.5    & 0.0          & 0.1          & 0.0         & 0.1        \\ \hline
\end{tabular}
\caption{Accuracy on the LAMBADA and bAbI benchmarks that we used to assess the robustness of \model to visual attacks generated following the procedure proposed by \citet{zero}.}
\label{tab:attack}
\end{table}

\end{document}